# An Identification System Using Eye Detection Based On Wavelets And Neural Networks

Mohamed A. El-Sayed
Dept. of Mathematics, Faculty of Science,
Fayoum University, Egypt
Assistant professor of CS, Taif University, KSA
msa06@fayoum.edu.eg

Mohamed A. Khfagy
Dept. of Mathematics, Faculty of Science,
Sohag University, Egypt
okalia2003@yahoo.com

*Abstract*—The randomness and uniqueness of human eye patterns is a major breakthrough in the search for quicker, easier and highly reliable forms of automatic human identification. It is being used extensively in security solutions. This includes access control to physical facilities, security systems and information databases, Suspect tracking, surveillance and intrusion detection and by various Intelligence agencies through out the world. We use the advantage of human eye uniqueness to identify people and approve its validity as a biometric. . Eye detection involves first extracting the eye from a digital face image, and then encoding the unique patterns of the eye in such a way that they can be compared with pre-registered eye patterns. The eye detection system consists of an automatic segmentation system that is based on the wavelet transform, and then the Wavelet analysis is used as a pre-processor for a back propagation neural network with conjugate gradient learning. The inputs to the neural network are the wavelet maxima neighborhood coefficients of face images at a particular scale. The output of the neural network is the classification of the input into an eye or non-eye region. An accuracy of 90% is observed for identifying test images under different conditions included in training stage.

I. INTRODUCTION

A biometric system provides automatic recognition of an individual based on some sort of unique feature or characteristic possessed by the individual. Biometric systems have been developed based on fingerprints, facial features, voice, hand geometry, handwriting, the retina, and the one presented in this research, the eye. Biometric systems work by first capturing a sample of the feature, such as recording a digital sound signal for voice recognition, or taking a digital color image for eye detection. The sample is then transformed using some sort of mathematical function into a biometric template. The biometric template will provide a normalized, efficient and highly discriminating representation of the feature, which can then be objectively compared with other templates in order to determine identity. Most biometric systems allow two modes of operation. A training mode or enrolment mode for adding templates to a database, and an identification mode, where a template is created for an individual and then a match is searched for in the database of pre-enrolled templates [1].

Recognition of human faces out of still images or image sequences is an actively developing research field. There are many different applications for systems coping with the problem of face localization and recognition e.g. model based video coding, face identification for security systems, gaze detection and human computer interaction. The detection and location of the face as well as the extraction of facial features from the images are crucial. Due to variations in illumination, background, visual angle and facial expressions, the problem is complex. In the first step of face recognition, the localization of facial regions within the facial contours is followed by the detection of facial features such as eyes, nose and mouth. Our algorithm for eye detection using wavelet transform and neural network is robust against changes in light conditions, visual angle, and noise in addition to being low in computational cost [2].

Yuille et al. [3] first proposed using deformable templates in locating human eye. The weaknesses of the deformable templates are that the processing time is lengthy and success relies on the initial position of the template.Lam et al. [4] introduced the concept of eye corners to improve the deformable template approach. Saber et al. [5] and Jeng et al. [6] proposed to use facial features geometrical structure to estimate the location of eyes. Takacs et al. [7] developed iconic filter banks for detecting facial landmarks.

The most common approach employed to achieve eye detection in real-time [8, 9, 10, 11] is by using infrared lighting to capture the physiological properties of eyes and an appearance-based model to represent the eye patterns. The appearance-based approach detects eyes based on the intensity distribution of the eyes by exploiting the differences in appearance of eyes from the rest of the face. This method requires a significant number of training data to enumerate all possible appearances of eyes i.e. representing the eyes of different subjects, under different face orientations, and different illumination conditions. The collected data is used to train a classifier such as a neural net or support vector machine to achieve detection. Various other methods that have been adopted for eye detection include wavelets, principal component analysis, fuzzy logic, support vector machines, neural networks, evolutionary computation and hidden markov models. Huang and Wechsler [12] perform the task of eye detection by using optimal wavelet packets for eye representation and radial basis functions for subsequent classification of facial areas into eye and non-eye regions. Filters based on Gabor wavelets to detect eyes in gray level images are used in [13]. Talmi et al. and Pentland et al. [14] use principal component analysis to describe and represent the general characteristics of human eyes with only very few dimensions. In [14] Eigeneyes are calculated by applying Karhunen-Loeve-Transformation to represent the major characteristics of human eyes and are stored as reference patterns for the localization of human eyes in video images. In this paper, we describe a novel algorithm for eye detection that is robust against changes in light conditions, visual angle, and noise in addition to being low in computational cost.

This paper is organized as follows: Section 2 presents some fundamental concepts and we describe the proposed method used. In Section 3, we report experiment discussion, the effectiveness of our method when applied to some real-world and some standard database set of images, such Olivetti Research Laboratory in Cambridge, UK [17], Japanese Female Facial Expression (JAFFE) database [18] and we also used non standard live images. At last Conclusion of this paper will be drawn in section 4.

## II. METHOD MATERIAL

The system consists mainly of two stages training and detection stage. A block diagram of these two stages is shown in Figure 1.

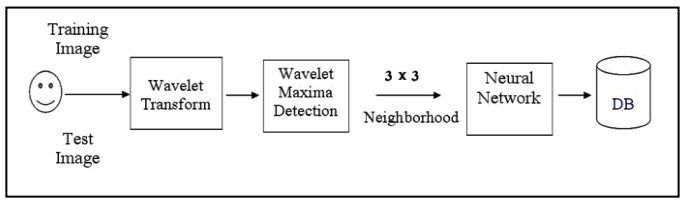

Figure 1. Stages of proposed method

### A. Acquisition of Training Data:

The training data typically consists of images of different persons with different hairstyles, different illumination conditions and varying facial expressions. Some of the images have different states of the eye such as eyes closed. The size of the images varies from 64x64 to 256x256.

### B. Wavelets

Wavelets are functions that satisfy certain mathematical requirements and are used in presenting data or other functions, similar to sines and cosines in the Fourier transform. However, it represents data at different scales or resolutions, which distinguish it from the Fourier, transform [15]. Wavelet decomposition provides local information in both space domain and frequency domain. Despite the equal subband sizes, different subbands carry different amounts of information. The letter 'L' stands for low frequency and the letter 'H' stands for high frequency. The left upper band is called LL band because it contains low frequency information in both the row and column directions. The LL band is a coarser approximation to the original image containing the overall information about the whole image. The LH subband is the result of applying the filter bank column wise and extracts the facial features very well. The HL subband, which is the result of applying the filter bank row wise, extracts the outline of the face boundary very well. While the HH band shows the high frequency component of the image in non-horizontal, non-vertical directions it proved to be a redundant subband and was not considered having significant information about the face. This observation was made at all resolutions of the image [16].

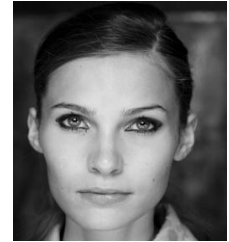

Figure 2. Original image, used to describe the proposed method

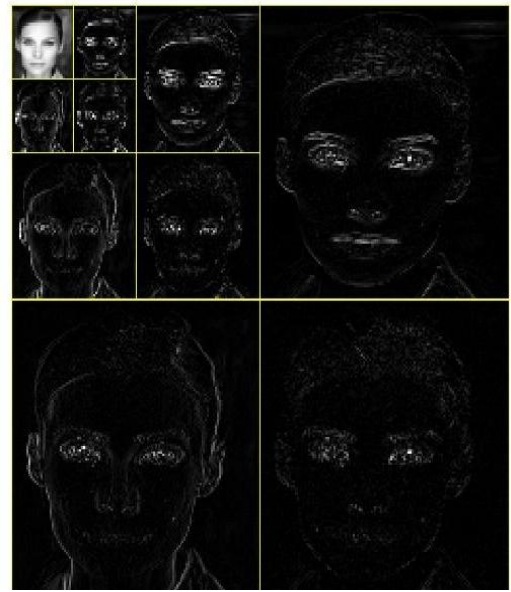

Figure 3. Discrete wavelet transform of original image

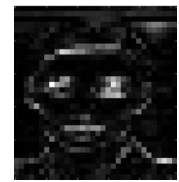

Figure 4. LH Subband of size 32×32

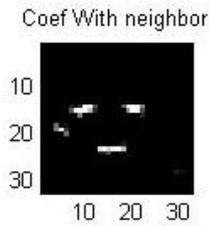

Figure 5. Wavelet Maxima' s

This is the first level decomposition. A CDF (2,2) bi-orthogonal wavelet is used. Gabor Wavelets seem to be the most probable candidate for feature extraction. But they suffer from certain limitations i.e. They cannot be implemented using Lifting Scheme and secondly the Gabor Wavelets form a non-orthogonal set thus making the computation of wavelet coefficients difficult and expensive. Special hardware is required to make the algorithm work in real time. Thus choosing a wavelet for eye detection depends on a lot of trial and error. Discrete Wavelet Transform is recursively applied to all the images in the training data set until the lowest frequency subband is of size 32×32 pixels i.e. the LH subband at a particular level or depth of DWT is of size 32×32. The original image's grayscale image is shown in Figure 2. The LH subband at resolution 32×32 is shown in Figure 3. Here we have used HAAR wavelet instead of Gabor wavelet while calculating wavelet transform.

We take the modulus of the wavelet coefficients in the LH subband. Experiments were performed to go to a resolution even coarser than 32×32. However, it was observed that in certain cases the features would be too close to each other and it was difficult even manually too to separate them. This would burden the Neural Network model and a small error in locating the eyes at this low resolution would result in a large error in locating the eyes in the original image.

*Detection of Wavelet Maxima:* Our approach to eye detection is based on the observation that, in intensity images eyes differ from the rest of the face because of their low intensity. Even if the eyes are closed, the darkness of the eye sockets is sufficient to extract the eye regions. These intensity peaks are well captured by the wavelet coefficients. Thus, wavelet coefficients have a high value at the coordinates surrounding the eyes. As shown in Figure 5.

C. *Neural Network 'Training*

The wavelet peaks detected are the center of potential eye windows. We then feed 3x3 neighborhood wavelet coefficients of each of these local maxima's in 32×32 LH subbands of all training images to a Neural Network for training. Here we have used the MLP (multi-layer perceptions) back-propagation model for neural network training. It consists of having 9 input nodes, 6 hidden nodes, and 2 output nodes. A diagram of the Neural Network architecture is shown in Figure 6. If we have a value of A (0, 1) at the output of Neural Network indicates an eye at the location of the wavelet maxima whereas (1, 0) indicates a non-eye. Two output nodes instead of one were taken to improve the performance of the Neural Network. MATLAB's Neural Network Toolbox was used for simulation of the back-propagation Neural Network. A conjugate gradient learning rate of 0.5 was chosen while training. This completes the training stages for neural networks back propagation model .

*Eye Localization*: Discrete Wavelet Transform is recursively applied to the test image until the lowest frequency subband is of size 32×32 pixels i.e. the LH subband is of size 32×32. The test image size maybe an integer multiple of 64×64. The absolute values of the wavelet coefficients are taken in this subband. Wavelet peaks in this subband are detected which are location of the potential eye windows. These peaks are then replaced with 3×3 neighborhood wavelet coefficients from the previous image and then fed to the Neural Network. The Neural Network classifies each of the peaks as eye or non-eye in this 32×32 subband.

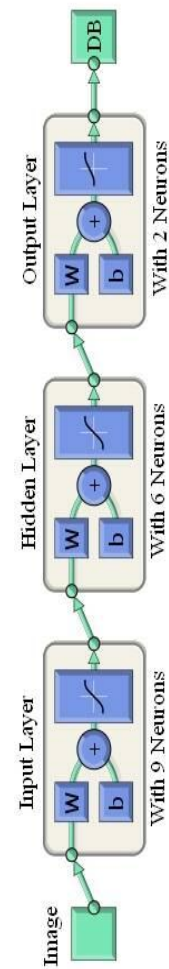

Figure 6. Neural Network Diagram

*D. Identity Extraction*

After training the network the updated weight and bias values for a particular person is stored in a database. The image to be verified is wavelet transformed before being applied to the neural network with those updated weight and bias values. The person is identified when the neural network output of one of the test images matches with that of the verified image as shown in Figure 7.

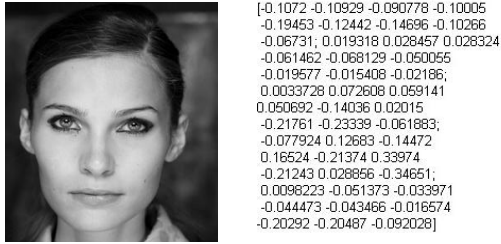

Figure 7. Identity Value For Training Image.

### III. EXPERIMENT RESULTS.

Eye detection is a pre-requisite stage for many applications such as the system proposed, template based eye detection is widely used recently, where an eye template is used to detect eye region from face image. The template is matched with eye region using cross correlation technique [19]. These methods are simple, does not require any complex mathematical calculation and prior knowledge about the eye, but they are not robust against illumination, background, facial expression changes and also works for images of different sizes, what makes the proposed method better to use.

In order to evaluate the proposed system a number of experiments were done to test the robustness of the algorithm and to increase the accuracy of eye detection. Various architectures of Neural Networks with different learning rates were tried and it was found that back propagation with conjugate gradient learning seemed to be the best choice. The training and test images we used was obtained from many resources such as Olivetti Research Laboratory in Cambridge, UK [17], Japanese Female Facial Expression (JAFFE) database[18] and we also used a non standard live images to approve our system robustness against different environmental conditions.

Identification systems the uses fingerprint and iris recognition is very expensive but here we use only a simple digital camera and processor to accomplish that proposed system that makes it easy can easily be implemented by hardware.

Any use of wavelets begins with Haar wavelet, the first and simplest. Haar wavelet is discontinuous, and resembles a step function. It represents the same wavelet as Daubechies db1. So Haar wavelet is the best choice for its simplicity and stability.

The network's performance according to the mean of squared errors (MSE) shows significant changes during training; hence it decreases gradually as shown in Figure 8.

Fletcher-Reeves conjugate gradient algorithm is used because it has smallest storage requirements of the conjugate gradient algorithms, that makes gradient shows a hard slope during changes as shown in Figure 9.

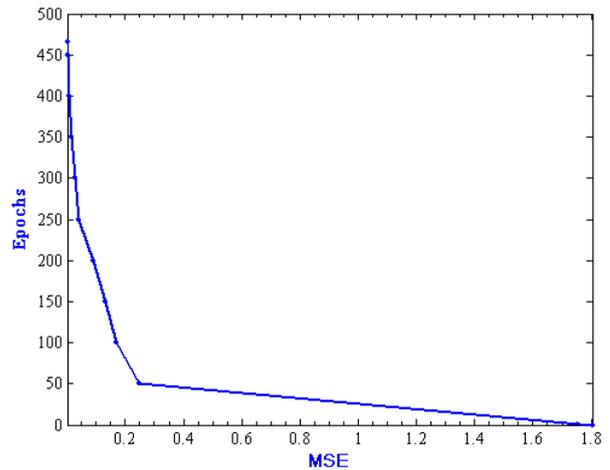

Figure 8. Network performance during training with MSE

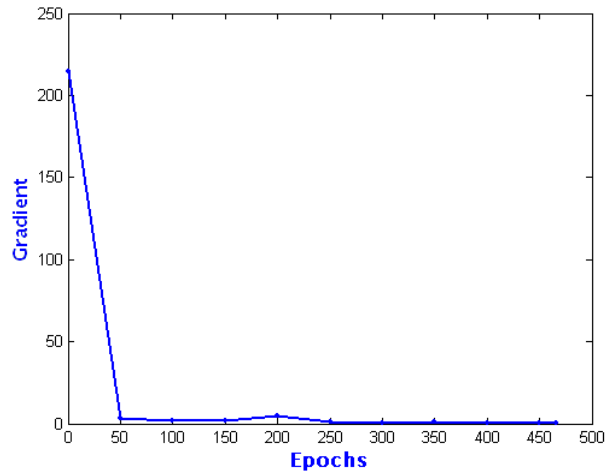

Figure 9. Gradient changes during training process.

TABLE I. MSE, ANN GRADIENT DURING TRAINING PROCESS.

| Epochs | MSE | ANN Gradient |
|--------|-----|--------------|
| 0 | 1.80372 | 214.684/1e-006 |
| 50 | 0.24862 | 3.16956/1e-006 |
| 100 | 0.168526 | 1.97275/1e-006 |
| 150 | 0.133475 | 1.67499/1e-006 |
| 200 | 0.0922139 | 4.63736/1e-006 |
| 250 | 0.0412001 | 1.20884/1e-006 |
| 300 | 0.026139 | 0.705481/1e-006 |
| 350 | 0.0173279 | 1.21542/1e-006 |
| 400 | 0.00531223 | 0.514081/1e-006 |
| 450 | 0.00154849 | 0.267079/1e-006 |
| 466 | 0.000993533 | 0.0892796/1e-006 |

The values of MSE and Gradient of ANN shows a similar behavior during training what reflects the system stability, table1 shows their values.

When the Performance goal is not met the network should be retrained with different parameters to reach the Performance goal, the learning rate had to be raised hence network shows complexity after training several inputs.

During the training process we tried to optimize the network parameters to suite it for the whole dataset specially the learning rate which we had to raise it gradually. The training process stops either when the performance goal reached as shown in figure 10. Or the error graph dose not shows any significant changes as shown in Figure 11.

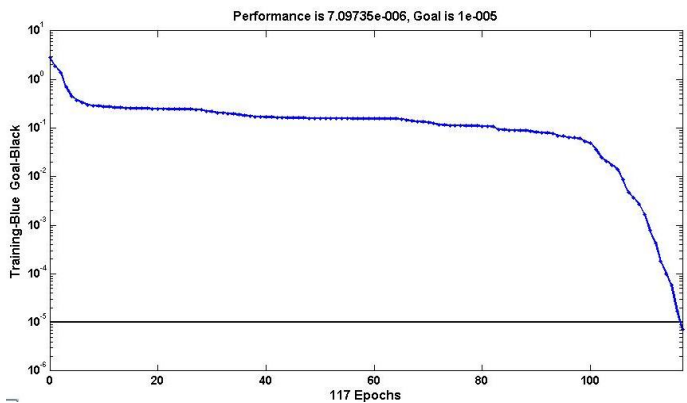

Figure 10. Conjugate Training Error Curve
Performance goal met after 117 epochs

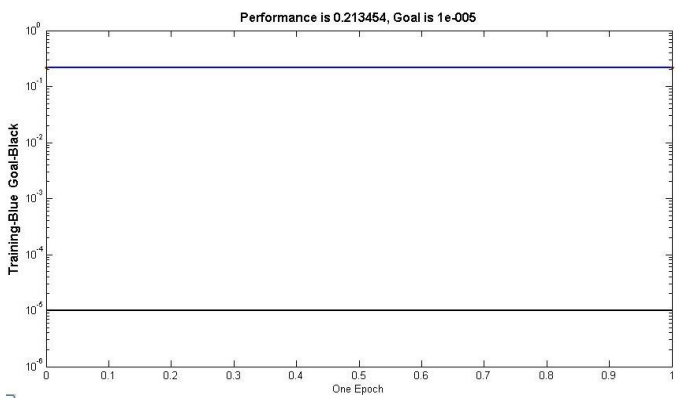

Figure 11. Conjugate Training Error Curve
Training Stopped After one epoch .

The Identification system performed well when applied on images used we could extract the image identity for the whole training images with an accuracy of 100% but the only drawback of it is to verify an image of the case of the images with closed eyes and images with complex background which is considered a special case with percentage about 90% so the system is very efficient to identify people, also its run time is very good for verification depends on the machine used.

IV. CONCLUSION.

This paper presents a new technique based on the use of wavelet and Neural Network to perform eye detection and apply it to an identification system. The present algorithm is robust and at par with the other existing methods but still has a lot of scope for improvement. In this type of approach a wavelet subbands approach in using Neural Networks for eye detection. Wavelet Transform is adopted to decompose an image into different subbands with different frequency components. A low frequency subband is selected for feature extraction. The proposed method is robust against illumination, background, facial expression changes and also works for images of different sizes. However, a combination of information in different frequency bands at different scales, or using multiple cues can even give better performance.

REFERENCES

[1] Igor Bőhm , Florian Testor"Biometric Systems" Department of Telecooperation University of Linz 4040 Linz, Austria.2006.

[2] S. Asteriadis, N. Nikolaidis, A. Hajdu, I. Pitas"An Eye Detection Algorithm Using Pixel to Edge Information"ISCCPSP 2006.

[3] A. L. Yuille, P. W. Hallinan, D. S. Cohen, "Feature extraction from faces using deformable templates", International Journal of Computer Vision 8(2) 99-111, (1992).

[4] K. M. Lam, H. Yan, " Locating and extracting the eye in human face images", Pattern Recognition, Vol. 29, 771-779, No. 5 (1996).

[5] E. Saber and A.M. Tekalp, "Frontal-view face detection and facial feature extraction using color, shape and symmetry based cost functions", Pattern Recognition Letters, Vol. 19, No. 8, pp. 669--680, 1998.

[6] Jeng, S.H., Liao, H.Y.M., Han, C.C., Chern, M.Y., Liu, Y.T., "Facial Feature Detection Using Geometrical Face Model: An Efficient Approach", Pattern Recognition, Vol. 31, No., pp. 273-282, 3, March 1998.

[7] Takacs, B., Wechsler, H., "Detection of faces and facial landmarks using iconic filter banks", Pattern Recognition, Vol. 30, No., pp. 1623-1636, 10, October 1997.

[8] X. Liu, F. Xu and K. Fujimura, "Real-Time Eye Detection and Tracking for Driver Observation Under Various Light Conditions", IEEE Intelligent Vehicle Symposium, Versailles, France, June 18-20, 2002.

[9] Zhiwei Zhu, Kikuo Fujimura, Qiang Ji, "Real-Time Eye Detection and Tracking Under Various Light Conditions and Face Orientations", ACM SIGCHI Symposium on Eye Tracking Research & Applications, New Orleans, LA, USA March 25th-27th 2002,.

[10] A. Haro, M. Flickner, and I. Essa, "Detecting and Tracking Eyes By Using Their Physiological Properties, Dynamics, and Appearance", Proceedings IEEE CVPR 2000, Hilton Head Island, South Carolina, June 2000.

[11] C. Morimoto, D. Koons, A. Amir, and M. Flickner, "Real-Time Detection of Eyes and Faces", Proceedings of 1998 Workshop on Perceptual User Interfaces, pages 117-120, San Francisco, CA, November 1998.

[12] J. Huang and H. Wechsler, "Eye Detection Using Optimal Wavelet Packets and Radial Basis Functions (RBFs)", International Journal of Pattern Recognition and Artificial Intelligence, Vol 13 No 7, 1999.

[13] S. A. Sirohey and Azriel Rosenfeld, "Eye detection in a face image using linear and nonlinear filters", Pattern Recognition. Vol 34., 1367-1391, 2001.

[14] K. Talmi, J. Liu, "Eye and Gaze Tracking for Visually Controlled Interactive Stereoscopic Displays", Signal Processing: Image Communication 14 799-810 Berlin, Germany (1999).

[15] Dao-Qing Dai and Hong Yan,Sun Yat-Sen (Zhongshan) " Wavelets and Face Recognition" Face Recognition, Book edited by: Kresimir Delac and Mislav Grgic, ISBN 978-3-902613-03-5, pp.558, I-Tech, Vienna, Austria, June 2007

[16] Hafiz Imtiaz and Shaikh Anowarul Fattah " A Face Recognition Scheme Using Waveletbased Dominant Features" Signal & Image Processing : An International Journal (SIPIJ) Vol.2, No.3, September 2011

[17] Olivetti Research Laboratory in Cambridge, UK" http://mambo.ucsc.edu /psl/olivetti.html".

[18] Japanese Female Facial Expression (JAFFE) Database" http://www. kasrl.org/jaffe.html".

[19] Nilamani Bhoi, Mihir Narayan Mohanty, " Template Matching based Eye Detection in Facial Image" International Journal of Computer Applications (0975   8887 Volume 12 No.5, December 2010